\documentclass[lettersize,journal]{IEEEtran}
\usepackage{amsmath,amsfonts}
\usepackage{algorithmic}
\usepackage{algorithm}
\usepackage{array}
\usepackage[caption=false,font=normalsize,labelfont=sf,textfont=sf]{subfig}
\usepackage{textcomp}
\usepackage{stfloats}
\usepackage{url}
\usepackage{verbatim}
\usepackage{graphicx}
\usepackage{cite}
\usepackage{multirow}
\usepackage{makecell}
\usepackage{diagbox}
\usepackage[table]{xcolor}
\usepackage[colorlinks,linkcolor=blue]{hyperref}
\begin{document}

\title{Real-World Depth Recovery via Structure Uncertainty Modeling and Inaccurate GT Depth Fitting}

\author{Delong Suzhang, Meng Yang,~\IEEEmembership{Member,~IEEE,}

\thanks{Manuscript received **, revised **. This work was supported by **.
}
}

\markboth{IEEE TRANSACTIONS ON JOURNAL NAME, MANUSCRIPT ID}%
{Shell \MakeLowercase{\textit{et al.}}: A Sample Article Using IEEEtran.cls for IEEE Journals}

\IEEEpubid{}

\maketitle
\begin{abstract}

The low-quality structure in raw depth maps is prevalent in real-world RGB-D datasets, which makes real-world depth recovery a critical task in recent years. However, the lack of paired raw-ground truth (raw-GT) data in the real world poses challenges for generalized depth recovery. Existing methods insufficiently consider the diversity of structure misalignment in raw depth maps, which leads to poor generalization in real-world depth recovery. Notably, random structure misalignments are not limited to raw depth data but also affect GT depth in real-world datasets. In the proposed method, we tackle the generalization problem from both input and output perspectives. For input, we enrich the diversity of structure misalignment in raw depth maps by designing a new raw depth generation pipeline, which helps the network avoid overfitting to a specific condition. Furthermore, a structure uncertainty module is designed to explicitly identify the misaligned structure for input raw depth maps to better generalize in unseen scenarios. Notably the well-trained depth foundation model (DFM) can help the structure uncertainty module estimate the structure uncertainty better. For output, a robust feature alignment module is designed to precisely align with the accurate structure of RGB images avoiding the interference of inaccurate GT depth. Extensive experiments on multiple datasets demonstrate the proposed method achieves competitive accuracy and generalization capabilities across various challenging raw depth maps.

\end{abstract}



\begin{figure*}[!t]
\centering
\subfloat[]{\includegraphics[width=3.5in]{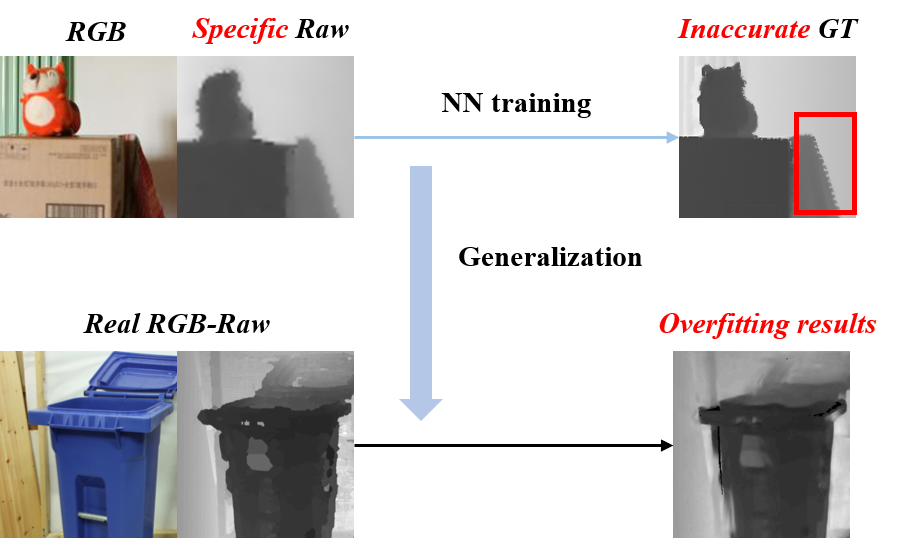}
\label{fig_first_case}}
\hfil
\subfloat[]{\includegraphics[width=3.5in]{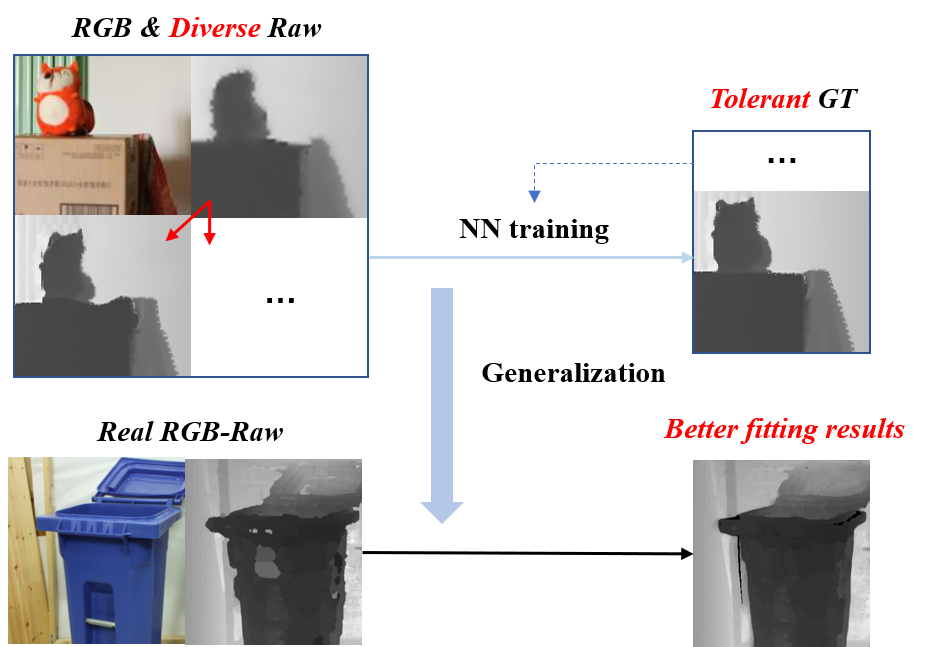}
\label{fig_second_case}}
\caption{The effects of real-world structure misalignment in depth maps. (a) conventional depth recovery training insufficiently accounts for the diversity of structural misalignment in real-world depth maps, leading to overfitting results during generalization tests. (b) our proposed approach explicitly models diverse structural misalignment patterns across both raw and GT depth maps, enabling robust generalization to real-world scenarios.}
\label{fig:1}
\end{figure*}

\section{Introduction}
Depth map conveys 3D scene information obtaining the object distance and the scene structure. It is widely utilized in various tasks, including semantic segmentation\cite{cong2024gradient}, 3D reconstruction \cite{gao2023dasi, RenDongRan}, object detection \cite{PRFANG2023109139, chun2024usd}, and robot localization \cite{agrawal2022dc}. With advancements in depth sensors\cite{nyuv2, ibims} or depth estimation methods \cite{li2018megadepth, midas}, obtaining depth information together with corresponding RGB images has become increasingly convenient. However, due to technical limitations, the structure quality of depth data is often lower than the corresponding RGB images. The low-quality depth in the real world severely affects downstream applications, making depth recovery critical research in recent years. 

A common idea for depth recovery is to leverage high-quality RGB information. Therefore, RGB-guided depth map recovery has gained significant attention over the past decade. Research can be broadly categorized into model-based methods and learning-based methods. Traditional model-based methods often formulate this task as constrained optimization problems, utilizing image filters or optimization models such as guided filter (GF) \cite{GuidedFilter}, Markov random fields (MRF) \cite{zuo2016mrf}, conditional random field (CRF) \cite{wang2023crf}. However, these methods are often time-consuming in solving optimization problems and are not always effective due to the limitation of artificially defined models. In recent years, the advent of deep learning has revolutionized this field, offering more flexible and powerful alternatives. To better utilize the precise structure of RGB, many insightful designs of neural networks have been proposed \cite{DKN2021, DAGF, wang2024sgnet}. They propose either an optimized convolutional module to mitigate the modality gap between RGB and depth, or a targeted attentional or diffusional model to improve training. However, these careful designs often struggle under the challenging real-world depth recovery \cite{he2021towards, wang2023crf}. 

One foundational bottleneck lies in that learning-based depth recovery rigidly demands the raw depth data and the ground truth (GT) depth data at the same time. Despite the rapid growth of depth-based applications and the extensive collection of RGB-D datasets, most datasets only provide either raw depth or GT depth, but not both. To address the issue, researchers have attempted to simulate raw depth \cite{DKN2021, DCTNet} or generate ideal GT depth \cite{yan2018ddrnet, wang2023self}. For example, major depth super-resolution methods \cite{DKN2021, DCTNet, dada, wang2024sgnet} employ bicubic interpolation to simulate low-resolution depth. However, models trained with such simulated data often perform poorly in real-world scenarios, such as realistic low-resolution Time-of-Flight (ToF) conditions \cite{he2021towards}, or under realistic heavy distortions \cite{wang2023crf}. As a result, these models frequently exhibit overfitting behavior in real-world tests, as illustrated in Fig. \ref{fig:1}.

Some researchers attribute the phenomenon to the gap between simulated and real-world data. Specifically, they captured paired raw-GT data in the real world \cite{he2021towards} or directly built virtual datasets and designed domain adaption in the real world \cite{qiao2023self, wang2023self}. However, these methods insufficiently consider the inherent diversity of realistic structure misalignments in depth maps. That is to say, the random structure misalignments could result in infinite realistic conditions in the real world, but not a fixed specific condition. However, learning from one specific condition repeatedly is one of the key reasons why networks exhibit overfitting behavior. Moreover, random structure misalignments are not only prevalent in various raw depth maps but also manifest as inaccuracies in the GT depth data, which also leads to a challenging problem for generalized depth recovery. 

In this paper, we recover various real-world raw depth maps in both input and output aspects. For input, we pioneer to enrich the diversity of structure misalignments in real-world raw depth maps by designing a new raw depth generation pipeline, avoiding the network overfitting to a specific condition. Besides the conventional noisy low-resolution degradation, the challenging structure misalignment in raw depth maps is considered through elastic transformation \cite{simard2003best, touvron2022resmlp}. Furthermore, a structure uncertainty module is designed to identify the misaligned structure in various simulated raw depth maps. The misaligned structure will be explicitly estimated and eliminated to mitigate the effects on final recovery results, which strongly helps generalization. Additionally, the depth foundation models \cite{midas, dav2} are developed into the proposed model as part of the input to better handle the uncertainty of various raw depth maps. For output, we design a robust feature alignment module to align output depth with the precise structure of RGB avoiding the interference of inaccurate GT depth. Inspired by the relationship between pixel-level features in RGB super-resolution \cite{liif}, we develop the module to align the precise structure of RGB with output depth in a learnable Multi-Layer Perception (MLP). The module is compatible with mainstream architectures (U-Net \cite{wang2024g2}, Vision Transformers (ViT) \cite{dav2}, and ConvNeXt \cite{woo2023convnext}) as a tail network to deal with high-dimensional features. Given the random distortions also exist in GT depth, the parameter update strategy of this module is further adopted as stochastic depth \cite{woo2023convnext} to avoid overfitting inaccurate GT depth. 

To validate the effectiveness of the proposed method, we conduct extensive experiments on widely used datasets, including RGBDD \cite{rgbdd} and Middlebury 2014 \cite{middlebury2014}. Our experiments cover three types of raw depth recovery tasks commonly encountered in real-world applications: (1) real-world low-quality depth captured by ToF sensors, (2) heavily distorted depth generated by a stereo matching algorithm \cite{sgbm}, and (3) noisy depth super-resolution recovery. For a fair comparison, the results are evaluated under two settings: generalization tests and in-domain tests, following \cite{wang2024sgnet, wang2023crf}. Both quantitative and qualitative results demonstrate that our method achieves superior performance compared to state-of-the-art baselines.

Our main contributions are highlighted as follows:
\begin{itemize}
    \item 
    We designed a raw depth generation pipeline to address the challenging structure misalignment in real-world raw depth maps. The data generation pipeline effectively enriches the diversity of structure misalignments and helps the networks avoid overfitting to a specific condition.

    \item 
    With the help of depth foundation model, we proposed a structure uncertainty module to explicitly identify the misaligned structure in raw depth maps, which strongly helps improve the generalization ability of depth recovery networks.
    
    \item 
    We designed a feature alignment module to align the output with the accurate structure of RGB while mitigating the interference of inaccurate GT depth. The module is compatible with common backbones including U-Net, ViT, and ConvNeXt to improve the quality of the recovered depth map.

    \item 
    Extensive experiments demonstrate the robustness and advancement of our total framework recovering multiple types of raw depth in the real world. 
\end{itemize}

\section{Related work}
\subsection{Model-based depth map recovery}
Over the past few decades, traditional model-based methods have been extensively studied for depth map recovery. Early methods primarily design specific filter kernels to weigh the pixels in spatial neighborhoods, thereby inferring new pixel values. The classical filter-based methods can recover depth maps with the help of RGB images, such as bilateral filter \cite{bf}, weighted mean filter \cite{zhang2014new}, and guided filter \cite{GuidedFilter}. Given the importance of depth map structure, more structure-preserving methods are explored. For instance, Ham et al. \cite{Ham2017RobustGIF} proposed static and dynamic guidance to simultaneously address the inconsistencies between the guidance images and the depth images. Liu et al. \cite{liu2021generalized} 
introduced truncated Huber penalty function to achieve the balance between smoothing and structure preservation. However, since local pixel inference often ignores global relationships in the depth map, methods based on global optimization have been proposed. Zuo et al. proposed an explicit edge inconsistency model to address the RGB-D correspondence in an MRF model \cite{zuo2016mrf} and a minimum-spanning forest model \cite{zuo2018minimum}. Liu et al. \cite{liu2016robust} developed the common L2 function into a robust penalty function to enhance depth maps. Wang et al. \cite{wang2022tip} investigated the inconsistency issues between low-quality depth maps and RGB images. Also, Wang et al. \cite{wang2023crf} redesigned data terms and smooth terms in dense CRF models to recover severely distorted depth maps.

Early model-based depth recovery methods leveraged specific filter kernels to enhance raw depth maps but ignored the global relationship in the depth map. Although global optimization models such as MRF and CRF have been utilized, they are constrained by artificially defined models and exhibit limitations in real-world depth recovery tasks.
    
\subsection{Learning-based depth map recovery}
In contrast to model-based methods, learning-based methods, which contain a large number of parameters, are designed to capture the underlying mapping relationships in large-scale RGB-D datasets. However, most general RGB-D datasets only provide either raw depth data or GT depth data, whereas supervised methods require both raw and GT depth data simultaneously. This data limitation has led learning-based depth recovery methods to be categorized into two types: conventional depth map recovery based on simulated paired data and generalized real-world depth map recovery.

{\textbf{Conventional depth map recovery based on simulated paired data. }} Due to the lack of paired raw-GT data, many learning-based recovery methods train and validate on simulated paired raw-GT data. Consequently, numerous remarkable network modules or training strategies have been proposed to effectively utilize high-quality RGB data. For example, Kim et al. \cite{DKN2021} developed a deformable convolution to address the depth enhancement problem. Zuo et al. \cite{zuo2019depth} improved the effectiveness of guidance features in a coarse-to-fine manner while accessing multi-scale guidance features. Zhao et al. \cite{DCTNet} employed discrete cosine transformation in neural networks for depth super-resolution. Metzger et al. \cite{dada} pioneered the use of anisotropic diffusion models for depth super-resolution. Zhong et al. \cite{DAGF} advanced attentional mechanisms to improve the effectiveness of guidance. Wang et al. \cite{wang2024sgnet} proposed a structure-guided network aiming at the gradient-frequency awareness for depth map super-resolution. In order to generate the raw-GT pairs for supervised training, these methods simulate the raw depth map insufficiently considering the random and diverse structure misalignments. For example, most depth super-resolution methods leverage the bicubic interpolation to simulate low-resolution depth maps. However, the gap between simulation and reality causes poor performance in real-world depth recovery. 

{\textbf{Generalized real-world depth map recovery. }} In real-world applications, recent research has increasingly focus on the gap between simulation and reality. Unlike conventional simulation-based methods, He et al. \cite{he2021towards} constructed RGBDD dataset, which simultaneously includes real-world raw depth and GT. They also proposed a dedicated network to enhance the real-world raw depth. Wang et al. \cite{wang2023crf} highlighted the challenges posed by severe distortions in recovering real-world raw depth maps. Also, Wang et al. \cite{wang2024g2} designed a unified network framework capable of addressing generalized depth enhancement and depth estimation. Qiao et al. \cite{qiao2023self} proposed a self-supervised network and designed a contrastive multiview pretraining scheme. Also, Wang et al. \cite{wang2023self} tackled the issue of inaccurate GT depth by developing a self-supervised network that learns the ideal RGB-D correspondences. Yan et al. \cite{yan2024learning} designed a complementary correlations module for better adaptions in real-world raw depth maps. 

Although various schemes have been designed to recover real-world raw depth maps, the existing methods still insufficiently consider the diversity of structure misalignments present in the real-world data. These random structure misalignments do not only affect the raw depth but also the GT depth. Consequently, random structure misalignments in real-world depth maps will introduce the ill-posed raw GT correspondence, leading to gaps in the generalized depth recovery methods. 

\subsection{Existing RGB-D Datasets}

The growing utilization of depth data has driven the development of numerous RGB-D datasets in recent years. Notable examples include NYUv2 \cite{nyuv2}, a widely-used indoor dataset captured using Kinect sensors across diverse scenes, and KITTI \cite{kitti}, the predominant outdoor dataset acquired through LiDAR technology. A fundamental assumption in RGB-D data processing is pixel-level structural alignment between RGB and depth modalities, where object boundaries are expected to align precisely across both representations. However, this assumption is frequently violated in practice due to structural misalignment caused by lightweight depth sensors, significantly limiting the applicability of widely-used datasets such as NYUv2 \cite{nyuv2} and Matterport3D \cite{matterport3d}. While high-quality datasets such as Ibims \cite{ibims} and vKITTI \cite{vkitti} have been developed using more precise sensors or virtual techniques, their creation involves substantial time and financial costs. Furthermore, models trained exclusively on high-quality data often fail to generalize to real-world scenarios where low-quality depth data predominates. These limitations highlight the critical importance of depth recovery for practical RGB-D applications. 

Supervised depth recovery requires paired raw-GT depth data captured simultaneously. Despite the rapid expansion of RGB-D datasets in recent years, most existing datasets provide either raw depth or GT depth, but rarely both. To the best of our knowledge, RGBDD \cite{rgbdd} remains the only publicly available RGB-D dataset that contains paired raw-GT depth data acquired from multiple depth sensors. This scarcity of paired data presents a significant challenge for developing generalized depth recovery methods. Unlike well-established tasks such as semantic segmentation \cite{kirillov2023segment} or depth estimation \cite{dav2}, which benefit from large-scale datasets, depth recovery struggles with limited paired data availability, hindering the training of robust models capable of generalizing across diverse real-world scenarios.

\section{Method}

\subsection{Raw Depth Generation Considering Random Structure Misalignment}
\label{sec:data}

\begin{figure}[!t]
\centering
\includegraphics[width=3.3in]{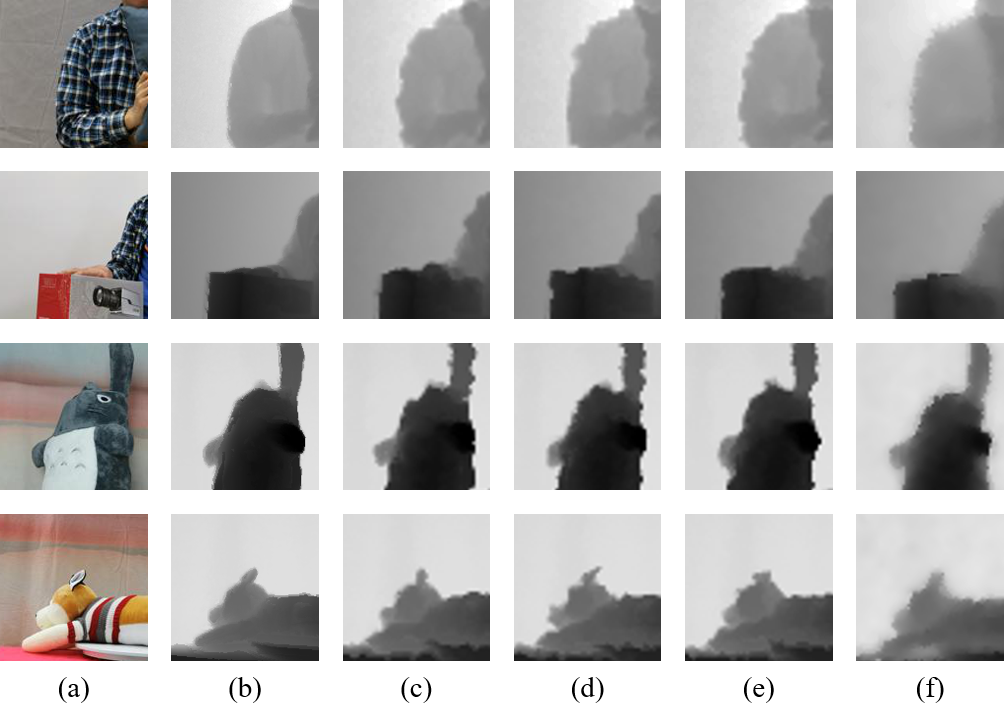}
\caption{The visual results of our raw depth generation. Beyond the conventional simulation of noisy super-resolution, we further simulate structural misalignment from GT depth maps, making the generated data more representative of real-world raw depth. (a) RGB image, (b) GT depth, (c), (d), and (e) represent different random structure misalignment simulations, (f) the real-world raw depth map captured by lightweight ToF sensor\cite{rgbdd}.}
\label{fig: data}
\end{figure}

Random structure misalignments are prevalent in raw depth maps, severely affecting the boundaries in raw depth maps captured by lightweight ToF sensor \cite{rgbdd}, as illustrated in Fig. \ref{fig: data}. The randomness of misalignment is obvious in visual results, while training on such fixed misalignment patterns will lead to the overfitting issue. In our solution, conventional noisy low-resolution degradation and structure misalignment degradation are both simulated in the proposed raw depth generation.

Our simulation focuses on the random generation of structural misalignment in raw depth maps. Here we adopt the random elastic transformation algorithm \cite{simard2003best, touvron2022resmlp}, which was originally designed to enrich the data of handwritten digits in terms of twisted deformation. 
As a result, we found it suitable for the simulation of structure misalignment. The algorithm of random elastic transformation is formulated as follows:

\begin{equation}
    D_{gen}=\Sigma_i sample(D^*;(x_i+dx_i, y_i+dy_i)),
\label{equ:ret}
\end{equation} where $sample$ in \eqref{equ:ret} denotes each value of a pixel is sampled by a new coordinate $(x_i+dx_i,y_i+dy_i)$ in $D^*$, whereas $D^*=\Sigma_i sample(D^*;(x_i,y_i))$. $dx_i$ and $dy_i$ denote the shift in coordinates in a similar pattern, as they are generated by follows:

\begin{equation}
    dx_i=sample(GB(\epsilon_x);(x_i,y_i)),
\end{equation}
\begin{equation}
    dy_i=sample(GB(\epsilon_y);(x_i,y_i)),
\end{equation} where $\epsilon_x\sim \mathcal{N}(0,1)$ and $\epsilon_y\sim \mathcal{N}(0,1)$ denotes the random Gaussian noise, $GB$ denotes the famous Gaussian Blur algorithm \cite{simard2003best}, with the parameter $\sigma$ is commonly set as 10.

In this way, the structure misalignment will be randomly simulated by different $\epsilon_x$ and $\epsilon_y$. The visual results of random structure misalignment simulation are illustrated in Fig. \ref{fig: data}. Results show such simulation well enriches the diversity of structure misalignment in raw depth maps, effectively avoiding either overfitting to one specific condition. Furthermore, a data simulation strategy also avoids capturing too many repeated scenes in the real world. To our knowledge, this work represents the first application of elastic transformation to simulate random structural misalignment in raw depth maps. More ablation studies of the raw depth generation pipeline can be found in Section \ref{ablation}. Given the unpredictable nature of distortions, which can occur at any location, we further develop a random rectangular mask to simulate localized distortions in raw depth maps. Random distortions are applied within the rectangular mask, while the regions outside the mask (i.e., the reverse mask) remain consistent with the GT depth map. 


\begin{figure}[!t]
\centering
\includegraphics[width=3.0in]{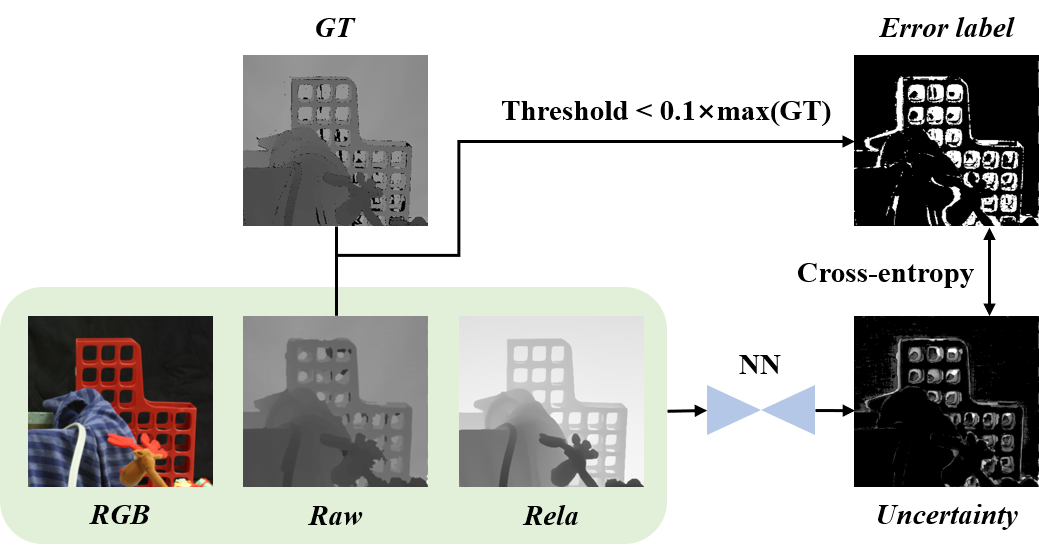}
\caption{The structure uncertainty estimation module of the raw depth map. A lightweight neural network (NN) is adopted to estimate the uncertainty of real-world raw depth maps with the classification solution. The error label is generated between raw depth and GT depth, with a threshold of $0.1\times max(GT)$ for binarization processing. }
\label{fig: uncertainty}
\end{figure}

\subsection{The Explicit Structure Uncertainty Estimation}
\label{sec:uncertainty}

Though structure misalignments are extensively simulated in raw depth maps as the input of neural networks, the factual misaligned structure will never be the same as one specific simulation. This means there is a gap between the simulated raw depth maps and the real raw depth maps. Furthermore, SOTA depth recovery methods \cite{DCTNet, wang2024g2} often maintain the misaligned structure in the generalized test, which also shows the lack of ability in structure identification. Therefore, a general structure uncertainty estimation module is proposed to mitigate the simulation gap and achieve better generalization. 

Specifically, the module employs a lightweight network architecture \cite{wang2024g2} to explicitly learn the structural uncertainty of various raw depth maps. The input and output details of this module are illustrated in Fig. \ref{fig: uncertainty}. The cross-entropy constraint is adopted to train the module as a classification problem, while the label of uncertainty is directly computed between the generated raw depth and the GT depth. The absolute difference less than threshold $\tau$ will be set as the label of 1 and set as 0 otherwise. Here threshold $\tau$ is set as $0.1*max(D^*)$. Furthermore, a similar binary mask based on the uncertainty map is set to the raw depth map, while the estimated wrong structure is set as 0 for the inputs instead of other values. Consequently, the uncertainty model explicitly helps address the various raw depth maps. The ablation study of the structure uncertainty model is shown in Section \ref{ablation}.

To mitigate the gap between RGB and depth, we further leverage the relative depth estimation generated by a depth foundational model \cite{dav2} to help address the structure uncertainty of raw depth. In contrast to conventional RGB-raw depth input, we concatenate the relative depth estimation results as an auxiliary channel, which mitigates the modality gap between RGB and raw depth. Note that relative depth estimation results may remain errors, thus we adopt it as the auxiliary input of a learnable network to better model the uncertainty of various raw depth maps. 

\subsection{Robust Feature Alignment Module}
\label{sec:GDF}
Random distortions also exist in real-world GT depth maps, affecting the optimization of the neural network. Although numerous remarkable designs have been proposed to align with the precise structure of RGB for raw depth recovery\cite{DKN2021, DAGF, wang2024sgnet}, random distortions in GT depth still result in improper guidance. In this subsection, we proposed a robust feature alignment module to align with the precise structure of high-quality RGB while mitigating the interference of inaccurate GT depth. The module is compatible with major network backbones, including U-Net \cite{unet}, Vision Transformer \cite{dav2}, and ConvNeXt \cite{woo2023convnext}. More details are illustrated in Fig. \ref{fig: framework}.

The module accepts the high dimensional RGB features $E(I) \in \mathbb R^{C \times H \times W}$ and the depth features $E(D_{r}) \in \mathbb R^{C \times H \times W}$ respectively extracted by major network backbones $E$ as input, where $C$, $H$, and $W$ denotes the channel dimension, height, and width of RGB-D features, respectively. Notice that the channel dimension (the number of hidden layers) $C$ is a hyper-parameter of the proposed module, with a default value of 64. We leverage separate $3\times3$ convolution to reduce the dimensions of RGB-D features. The process can be expressed as follows: 
\begin{equation}
    E'(I) =conv_{3 \times 3}(E(I)),
    E'(D_{r}) =conv_{3 \times 3}(E(D_{r})),
\end{equation} where $E'(C) \in \mathbb R^{1 \times H \times W}$ and $E'(D) \in \mathbb R^{1 \times H \times W}$ denotes the dimensionality reduction feature of RGB and depth, respectively.


To align with the pixel-level feature of precise RGB, we follow \cite{liif} and leverage coordinate encoding in the proposed module, developing the high-dimensional feature of RGB as relationship guidance for depth recovery. For each target pixel $i$, we define four query pixels $q_1, q_2, q_3, q_4$ located at the top-left, top-left, bottom-left, and bottom-right to help infer the feature value. The distance $f$ between the query pixel $q_j$ and the target pixel is set as 5. For each query pixel, the RGB feature $E(I)$ is sampled as $E_{q_{j}}(I) \in \mathbb R^{C \times H \times W}$. In order to better propagate gradients in coordinate encoding, the coordinate relationship is concatenated as a part of the input. The coordinate relationship between the query pixel and the target pixel along the x and y directions is denoted as $X_{q_{j},i}\in \mathbb R^{1 \times H \times W} $, $Y_{q_{j},i} \in \mathbb R^{1 \times H \times W}$, respectively. Then the sampled feature with its corresponding relative coordinate is concatenated together as $E_{q_{j},c}(I)  \in \mathbb R^{(C+2) \times H \times W}$, where $E_{q_{j},c}(I)=concat(E_{q_{j}}(I), X_{q_{j},i}, Y_{q_{j},i})$. After concatenating the relative coordinates with the sampled feature, $E_{q_{j},c}(I)$ is fed into a Multilayer Perception (MLP), and each query feature after MLP is stacked together to unify the relationship between query and target. The process can be expressed as follows: 
\begin{equation}
    a_{q_{j},i}, b_{q_{j},i} = MLP(E_{q_{j},c}(I)),
\end{equation} where $a_{q_{j},i}, b_{q_{j},i} \in \mathbb R^{ H \times W }$ denotes the weight and the residual between the query feature and the target feature, respectively. We adopt the standardized weight $\omega_{q_{j},i}=softmax(a_{q_{j},i})$ as the final weight in an attention manner to help convergence. To avoid overfitting the random distortion in GT depth map, we additionally adopt stochastic depth \cite{woo2023convnext} as the strategy of parameter optimization. The strategy can also be replaced as others, for example, dropout\cite{dropout}, dropblock\cite{dropblock}, and so on. The relevant ablation study can be found in Section \ref{ablation}. 

As a result, the output depth map is defined as follows:
\begin{equation}
    D = E'(I)+E'(D_{r})+\Sigma_{j}(\omega_{q_{j},i} \times b_{q_{j},i}),
\end{equation} where $D \in \mathbb R^{1\times H \times W}$ denotes the output of the network.

\subsection{Total framework and loss functions}

\begin{figure*}[!t]
\centering
\includegraphics[width=6in]{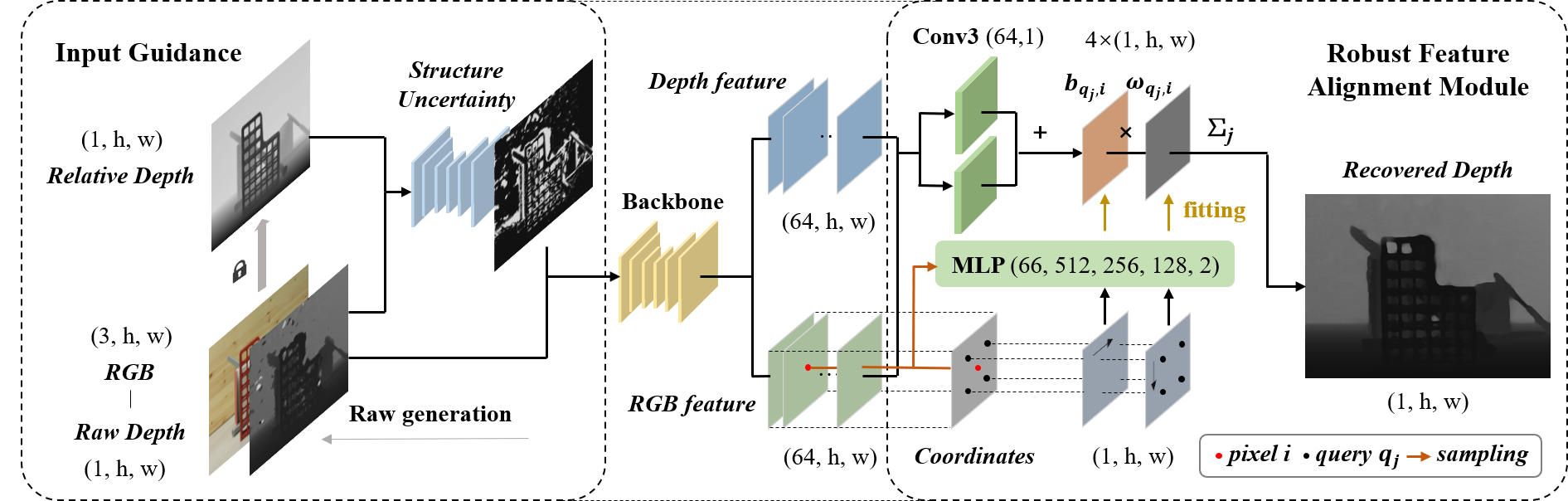}
\caption{The overall framework of our proposed model. The input and output of our robust feature alignment module are illustrated in detail. Specifically, the structure uncertainty model leverages a relative depth estimation from a depth foundation model (DFM) \cite{dav2} to quantify the uncertainty of various raw depth maps, with weights of DFM kept fixed during training. Additionally, our feature alignment module is designed to align with the precise RGB feature mitigating the interference of inaccurate GT depth. The module is compatible with different network backbones, including U-Net \cite{wang2024g2}, ViT \cite{dav2}, and ConvNeXt \cite{woo2023convnext}. }
\label{fig: framework}
\end{figure*}

The overall framework of our proposed model is illustrated in {Fig.\ \ref{fig: framework}}. It consists of three main components: (1) the raw depth generation pipeline in Section \ref{sec:data}, (2) the structure uncertainty model introduced in Section \ref{sec:uncertainty}, and (3) the depth recovery network incorporating the feature alignment module described in Section \ref{sec:GDF}. The training loss functions of depth recovery network consist of three conventional components: the L1 constraint \cite{wang2024g2}, the relative constraint \cite{midas, wang2024g2}, and the multi-scale gradient constraint \cite{midas, li2018megadepth}. The L1 constraint is defined as follows:

\begin{equation}
\label{equ7}
L_1(D,D^*)=\frac{1}{n}\sum_{i=1}^n|D_i-D_i^*|,
\end{equation} where the variable $n$ denotes the number of valid pixels in GT depth.

The relative constraint is utilized to learn the structure of the depth map robustly \cite{midas, wang2024g2}. The definition is as follows:

\begin{equation}
\label{equ8}
\begin{aligned}L_{r}(D,D^*)=\frac{1}{n}\sum_{i=1}^n\left|\frac{D_i-\overline{D}}{\sigma_D}-\frac{D_i^*-\overline{D^*}}{\sigma_D^*}\right|,\end{aligned}
\end{equation} \noindent where $\overline{D}$ and $\overline{D^*}$ denote mean values of output depth $D$ and GT depth $D^*$. $\sigma_D$ and $\sigma_D^*$ denote the standard deviations of output depth and GT depth, separately. 

Last, a common multi-scale gradient loss to constrain the structures between output depth and GT depth at different scales \cite{midas, li2018megadepth}. The definition is as follows:

\begin{equation}
\label{equ9}
\begin{aligned}L_{msg}(D,D^*)=\frac{1}{4n}\sum_{k=1}^4\sum_{i=1}^n\left|{\nabla_x}R_i^k-{\nabla_y}R_i^k\right|,\end{aligned}
\end{equation}

\noindent where $R_i=D_i-D_i^*$ denotes the residual between output depth and GT depth, $k$ denotes the $k$-th down-sampling scale, $\nabla_x$ and $\nabla_y$ denote the operators in direction of x and y, respectively.

The total loss functions are defined as follows:

\begin{equation}
\label{equ10}
\begin{aligned}
&L(\theta)=L_1(D,D^*)+L_{r}(D,D^*)+\lambda L_{msg}(D,D^*),
\end{aligned}
\end{equation} where $\theta$ denotes the parameter of networks, $\lambda$ is a hyper-parameter which is commonly set as 0.5 \cite{midas, li2018megadepth}. 

\section{Experiments}
\subsection{Dataset and Metric}
To evaluate the performance of our method, we conduct extensive experiments on widely-used datasets RGBDD \cite{he2021towards} and Middlebury 2014 \cite{middlebury2014}. Our experiments cover three types of raw depth recovery tasks commonly encountered in real-world applications: (1) low-quality depth captured by ToF sensors \cite{rgbdd}, (2) heavily distorted depth generated by a stereo matching algorithm \cite{sgbm}, and (3) simulated low-resolution noisy depth \cite{wang2022tip, wang2023crf}.  For the fair comparison, the results are evaluated set under two settings: in-domain test \cite{DCTNet, wang2024sgnet} and generalization test \cite{wang2024g2, wang2023crf}. 

We separately provide in-domain model and generalized model for in-domain test and generalization test, respectively. A benchmark for in-domain tests is established on the RGBDD dataset, where we train in-domain model following the same settings of \cite{DCTNet, wang2024sgnet}. For the generalized test, we also provide a model trained on mixed RGB-D datasets, including Matterport3D \cite{matterport3d}, HRWSI\cite{xian2020structure}, vKITTI \cite{vkitti}, and UnrealCV \cite{wang2024g2}. This model is applied in all generalized tests in this paper, which means the testing scenes are never seen during training.

We comprehensively use four widely-used metrics \cite{hu2022deep} to evaluate the quantitative results of our model and the baselines including absolute relative error (AbsRel), root mean square error (RMSE), root mean square error of inverse depth (iRMSE), the correct proportion of pixels in GT depth ($\delta_{1.05}$). The expressions of four metrics are as follows:
\begin{enumerate}
\item{AbsRel: $\frac{1}{n}\sum_{i=1}^{n}{|D_{i}-D^*_{i}|}/{D^*_{i}},$}
\item{RMSE: $\sqrt{\frac{1}{n}\sum_{i=1}^{n}(D_{i}-D^*_{i})^{2}},$}
\item{iRMSE: $\sqrt{\frac{1}{n}\sum_{i=1}^{n}({1}/{D_{i}}-{1}/{D^*_{i}})^{2}},$}
\item{$\delta_{1.05}$: $max(D_i/D^*_i, D^*_i/D_i)=\delta<1.05.$}
\end{enumerate} where $D$ and $D^*$ mean the output depth and GT depth, respectively.

We compare our model with publicly released codes provided by recent competitive baselines. For the in-domain test, all models are retrained on the target dataset, such as the RGBDD dataset, to ensure a fair comparison. For the generalization test, we evaluate baseline models that have not been trained on the target dataset, such as the Middlebury 2014 dataset. 

\subsection{Implementation Details}

Besides the random structure misalignments simulated in Section \ref{sec:data}, we also address the conventional noise and low-resolution degradation in raw depth maps. We employ a combination of random Gaussian noise and salt and pepper noise following \cite{wang2024g2}. For low-resolution degradation, we randomly select common interpolation algorithms, including bicubic, bilinear, and nearest-neighbor, in order to avoid the overfitting problem. The random resize rate is randomly sampled from a uniform distribution within the range [4, 16]. 

For the training of neural network, we totally trained two models for comparison: an in-domain model (IM) and a generalization model (GM). Both models are optimized using the Adam optimizer with a learning rate of 0.001 and a decay factor of 0.5 applied every 20 epochs. The IM is trained for 20 epochs, while the GM is trained for 40 epochs. The batch size for the IM is set to 1, whereas for the GM, it is set to 14. Both models are trained on two RTX 3090 GPUs. During training, we employ random cropping to generate RGB-D pairs with a resolution of 320 $\times$ 320. Any data with a resolution lower than 320 $\times$ 320 is excluded. To accelerate training, we utilize mixed precision training provided by PyTorch.

\subsection{Performance Comparison}
\subsubsection{Recovery of raw depth captured by ToF sensors}
The RGBDD dataset provides both distorted raw depth and GT depth captured by ToF sensors. Following \cite{DCTNet, wang2024sgnet}, we conduct both in-domain and generalization tests on this dataset.

The RGBDD dataset includes a benchmark for the in-domain test, where deep learning models can be trained and validated on this dataset. The quantitative results are presented in Table \ref{tab:rgbdd_in}. For this test, all models are trained within the official training split of RGBDD. Note that the results for DCT and SGN are obtained using their shared pre-trained weights and code. Since other learning-based baselines do not provide pre-trained weights, we retrained their publicly available models according to their original settings, denoted by an asterisk “*" in Table \ref{tab:rgbdd_in}. The results demonstrate that our method comprehensively achieves SOTA performance in the in-domain test, with a notable improvement of 18.75\% in AbsRel.

Moreover, the quantitative results of the generalization test on the RGBDD dataset are presented in Table \ref{tab:rgbdd_out}. Our method outperforms recent SOTA depth recovery methods in the generalization test on RGBDD. Notably, although C2F is a model-based method, it still demonstrates competitive performance compared to SOTA learning-based methods. As a learning-based approach, our method effectively addresses the diversity of distortions, which helps our model achieve SOTA performance in the generalization test.

\begin{figure*}[!t]
\centering
\includegraphics[width=6in]{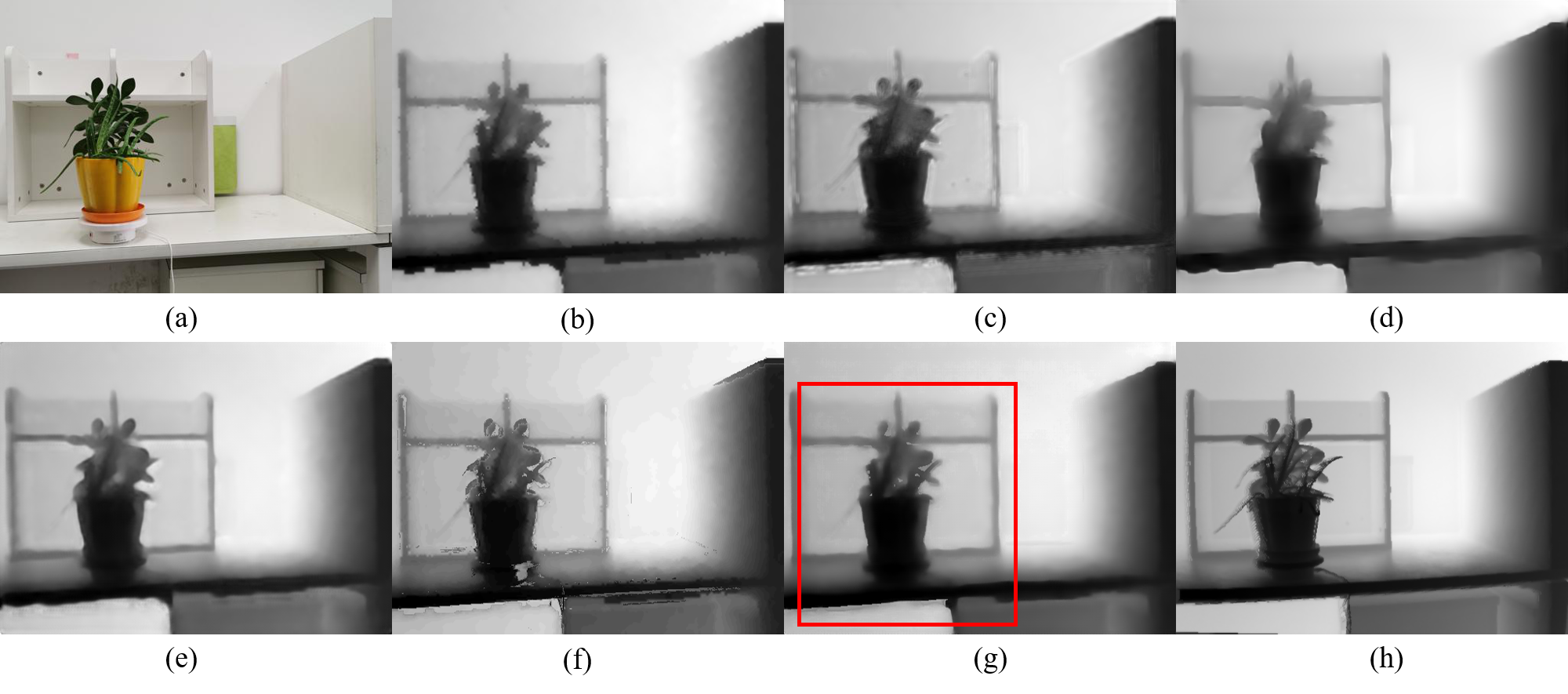}
\caption{Visual results of generalization test on RGBDD. (a) RGB, (b) Raw, (c) DCT\cite{DCTNet}, (d) G2\cite{wang2024g2}, (e) SGN\cite{wang2024sgnet}, (f) C2F \cite{wang2023crf}, (g) Ours, (h) GT.}
\label{fig:rgbdd_out}
\end{figure*}

The visual results of the generalization test on the RGBDD dataset are illustrated in Fig. \ref{fig:rgbdd_out}. Random structure misalignments are evident in the raw depth, and SOTA methods exhibit distinct characteristics when processing such data. For instance, the SOTA method DCT demonstrates the ability to locate misaligned structures but fails to reconstruct them accurately. G2 and SGN show smooth reconstruction capabilities but struggle to correct erroneous structures present in the raw depth. C2F recovers sharp edges but lacks effective smoothing. In contrast, our method achieves a balanced performance, effectively recovering structures while maintaining smoothness in the visual results.

\begin{table}[t]
\centering
\caption{The quantitative results comparison on the in-domain test on RGBDD. “*" denotes the model is retrained in RGBDD by us as their settings. The best are in \textbf{bold} and the second best are \underline{underlined} ones.}
\begin{tabular}{l@{ }c@{ }c@{ }c@{ }c@{ }c}
\hline
Method & reference & AbsRel$\downarrow$ & RMSE$\downarrow$ & iRMSE$\downarrow$ & $\delta_{1.05}\uparrow$ \\
\hline
DKN*\cite{DKN2021} & IJCV2021 & 0.023 & 7.649 & 2.876 & 91.5\\
DCT\cite{DCTNet} & CVPR2022 & 0.018 & 5.408 & \textbf{1.804} & 94.7\\
DADA*\cite{dada} & CVPR2023 & 0.022 & 7.290 & 2.555 & 91.9\\
SGN\cite{wang2024sgnet} & AAAI2024 & 0.017 & \underline{5.324} & 1.903 & 94.6\\
G2*\cite{wang2024g2} & PAMI2024 & \underline{0.016} & 5.388 & 1.920 & \underline{95.8}\\
Ours & - & \textbf{0.013} & \textbf{5.109} & \underline{1.836} & \textbf{96.0}\\
\hline
\end{tabular}
\label{tab:rgbdd_in}
\end{table}

\begin{table}[t]
\centering
\caption{The quantitative results comparison on the generalization test on RGBDD. The best are in \textbf{bold} and the second best are \underline{underlined} ones.}
\begin{tabular}{l@{ }c@{ }c@{ }c@{ }c@{ }c}
\hline
Method & reference & AbsRel$\downarrow$ & RMSE$\downarrow$ & iRMSE$\downarrow$ & $\delta_{1.05}\uparrow$ \\
\hline
DKN\cite{DKN2021} & IJCV2021 & 0.023 & 8.225 & 3.596 & 91.7\\
DCT\cite{DCTNet} & CVPR2022 & \underline{0.022} & 7.369 & 2.563 & 92.1\\
DADA\cite{dada} & CVPR2023 & \underline{0.022} & 7.320 & 2.560 & 92.1\\
C2F\cite{wang2023crf} & TIP2023 & \textbf{0.020} & \underline{6.873} & \underline{2.439} & \underline{92.8}\\
SGN\cite{wang2024sgnet} & AAAI2024 & \underline{0.022} & 7.213 & 2.528 & 92.0\\
G2\cite{wang2024g2} & PAMI2024 & \underline{0.022} & 7.174 & 2.535 & 91.8\\
Ours & - & \textbf{0.020} & \textbf{6.652} & \textbf{2.333} & \textbf{93.7}\\
\hline
\end{tabular}
\label{tab:rgbdd_out}
\end{table}

\subsubsection{Recovery of raw depth with heavy distortions}
The generalization ability of our model is further evaluated by recovering heavily distorted depth maps on the Middlebury 2014 dataset. Note that Middlebury 2014 does not provide raw depth maps and GT depth maps simultaneously but instead offers binocular RGB images and corresponding camera parameters for stereo matching. Following \cite{wang2022tip, wang2023crf}, we employ the Semi-Global Block Matching (SGBM) algorithm \cite{sgbm} to generate raw depth maps with heavy distortions \cite{wang2022tip, wang2023crf}. For the fair comparison, all depth recovery methods are evaluated under the generalization test setting, meaning the testing RGB-raw pairs are never seen during training.

\begin{figure*}[!t]
\centering
\includegraphics[width=6in]{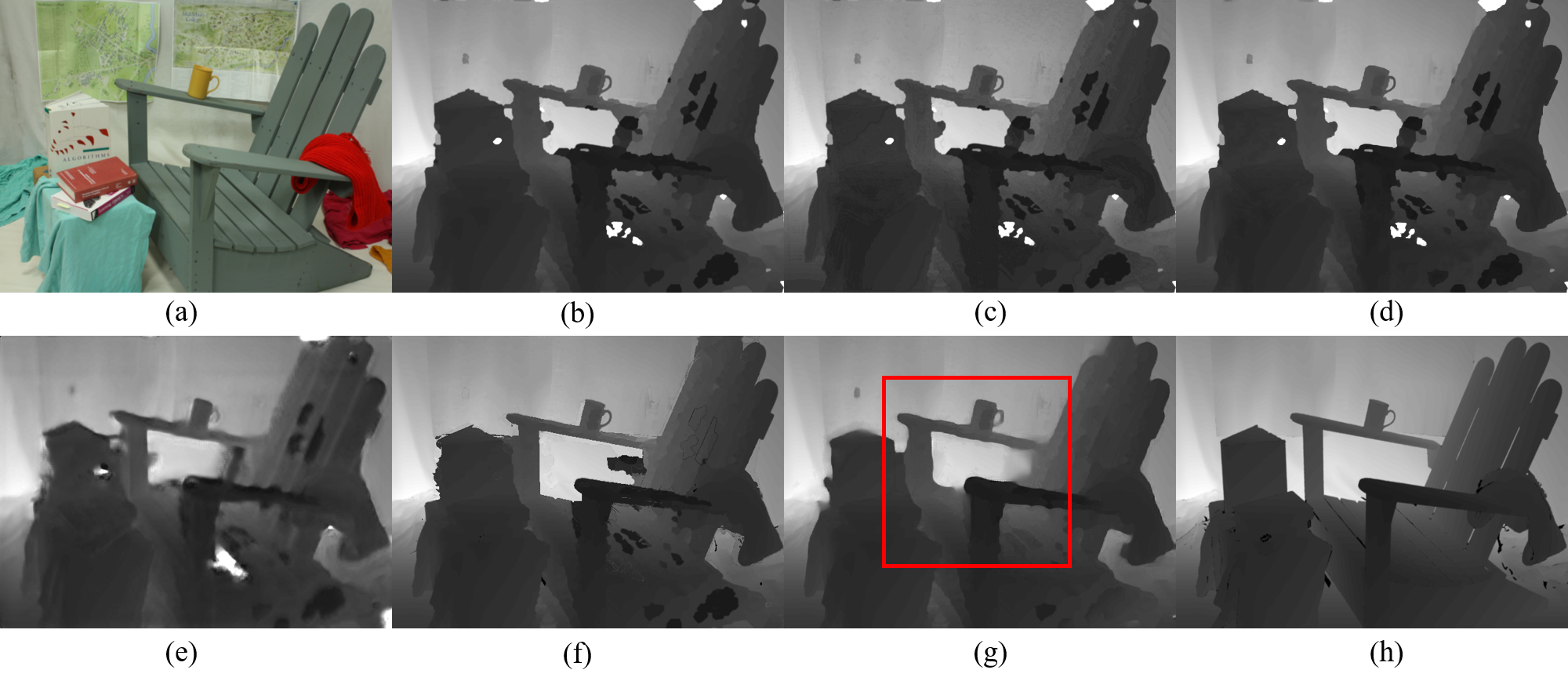}
\caption{Visual results of generalization test on Middlebury 2014. (a) RGB, (b) SGBM\cite{sgbm}, (c) DCT\cite{DCTNet}, (d) G2\cite{wang2024g2}, (e) SGN\cite{wang2024sgnet}, (f) C2F \cite{wang2023crf}, (g) Ours, (h) GT.}
\label{fig:mdb2014_out}
\end{figure*}

The quantitative results are presented in Table \ref{tab:mdb2014}, where our method achieves comprehensively the best accuracy among SOTA depth recovery baselines. Visual results are illustrated in Fig. \ref{fig:mdb2014_out}. Recent learning-based baselines, such as DCT and G2, tend to preserve incorrect structures from the raw depth maps, exhibiting overfitting characteristics. The SOTA baseline SGN demonstrates generalization capabilities but struggles with underfitting issues. Model-based method C2F is designed to recover the heavy distortions in a coarse-to-fine manner but still suffers from the predefined models and the wrong structure. In contrast, our method achieves SOTA performance in visual results while effectively handling unseen heavy distortions in raw depth maps. 

The generalization ability of our model benefits from our specific design in both input and output perspectives. For input, our raw depth generation pipeline effectively helps the model avoid overfitting a specific raw pattern. The structure uncertainty module improves the ability of identifying wrong structures. For output, the feature alignment module utilizes the high-quality RGB feature and helps not overfit to inaccurate GT depth. Therefore, the results of generalization test well proves the effectiveness and advancement of the proposed framework.

\begin{table}[t]
\centering
\caption{The quantitative results comparison on the generalization test of Middlebury 2014. The best are in \textbf{bold} and the second best are \underline{underlined} ones.}
\begin{tabular}{l@{ }c@{ }c@{ }c@{ }c@{ }c}
\hline
Method & reference & AbsRel$\downarrow$ & RMSE$\downarrow$ & iRMSE$\downarrow$ & $\delta_{1.05}\uparrow$ \\
\hline
DKN\cite{DKN2021} & IJCV2021 & 0.099 & 13.826 & 11.257 & 66.8\\
DCT\cite{DCTNet} & CVPR2022 & 0.095 & 13.067 & 2.719 & 62.5\\
DADA\cite{dada} & CVPR2023 & 0.093 & 13.184 & 3.645 & 68.2\\
C2F\cite{wang2023crf} & TIP2023 & \underline{0.044} & \underline{7.276} & \underline{1.930} & \textbf{84.6}\\
G2\cite{wang2024g2} & PAMI2024 & 0.085 & 13.153 & 2.602 & 73.5\\
SGN\cite{wang2024sgnet} & AAAI2024 & 0.123 & 12.483 & 2.678 & 55.6\\
Ours & - & \textbf{0.042} & \textbf{6.841} & \textbf{1.853} & \underline{81.5}\\
\hline
\end{tabular}
\label{tab:mdb2014}
\end{table}

\subsubsection{Results of noisy depth super-resolution}

The experiments above validate the ability of our method under noise-free conditions. We further implement the proposed method in the task of noisy depth super-resolution. Noisy low-resolution depth maps are generated following the experimental settings of \cite{wang2022tip, wang2023crf} for the RGBDD dataset and the Middlebury 2014 dataset, respectively. Notably, the noisy low-resolution depth maps of RGBDD are derived from the raw depth maps captured by the ToF tensor, while those of Middlebury 2014 are generated from results of SGBM\cite{sgbm}. For simplicity, we adopt the generalization models of all methods for validation. We report the same metrics of AbsRel, RMSE, iRMSE, and $\delta_{1.05}$ for 4$\times$, 8$\times$, and 16$\times$ depth super-resolution. The quantitative results for noisy depth super-resolution on RGBDD dataset are presented in Table \ref{tab:nsr_rgbdd}, and those on the Middlebury 2014 dataset are presented in Table \ref{tab:nsr_mdb2014}. The results in Table \ref{tab:nsr_rgbdd} and Table \ref{tab:nsr_mdb2014} demonstrate that our method comprehensively achieves the best accuracy across all scaling factors in both RGBDD and Middlebury 2014. Our method not only performs well in the noise-free condition but also in noisy depth super-resolution. 

\begin{table*}[t]
\centering
\caption{The quantitative results comparison of the noisy depth super-resolution test on RGBDD. The best are in \textbf{bold} and the second best are \underline{underlined} ones.}
\begin{tabular}{l@{ }c@{ }|c@{ }c@{ }c@{ }c@{ }|c@{ }c@{ }c@{ }c@{ }|c@{ }c@{ }c@{ }c}
\hline
\multirow{2}{*}{Method} & \multirow{2}{*}{reference} & \multicolumn{4}{c|}{4$\times$} & \multicolumn{4}{c|}{8$\times$} & \multicolumn{4}{c}{16$\times$}\\
 & & AbsRel$\downarrow$ & RMSE$\downarrow$ & iRMSE$\downarrow$ & $\delta_{1.05}\uparrow$ & AbsRel$\downarrow$ & RMSE$\downarrow$ & iRMSE$\downarrow$ & $\delta_{1.05}\uparrow$ & AbsRel$\downarrow$ & RMSE$\downarrow$ & iRMSE$\downarrow$ & $\delta_{1.05}\uparrow$ \\
\hline
DKN\cite{DKN2021} & IJCV2021 & 0.033 & 9.751 & 3.968 & 81.6 & 0.033 & 9.680 & 3.948 & 81.8 & 0.032 & 9.116 & 3.154 & 82.9\\
DCT\cite{DCTNet} & CVPR2022 & 0.037 & 10.340 & 3.780 & 78.4 & 0.030 & 8.370 & 3.004 & 84.5 & 0.031 & 8.581 & \textbf{3.036} & 83.5\\
DADA\cite{dada} & CVPR2023 & 0.033 & 9.268 & 3.371 & 81.2 & 0.034 & 9.440 & 3.459 & 80.6 & 0.035 & 9.852 & 4.352 & 79.4\\
C2F\cite{wang2023crf} & TIP2023 & \underline{0.028} & 8.192 & 2.953 & 85.2 & {0.029} & 8.240 & 2.996 & 84.9 & \underline{0.030} & \underline{8.499} & 3.102 & 83.3\\
G2\cite{wang2024g2} & PAMI2024 & \textbf{0.022} & \underline{6.802} & \underline{2.418} & \underline{91.9} & \textbf{0.023} & \underline{6.914} & \underline{2.419} & \underline{91.2} & \textbf{0.029} & 9.078 & 3.071 & \textbf{85.8}\\
SGN\cite{wang2024sgnet} & AAAI2024 & 0.038 & 10.640 & 3.905 & 76.8 & 0.030 & 8.311 & 2.990 & 84.7 & 0.031 & 8.551 & \underline{3.044} & 83.4\\
Ours & - & \textbf{0.022} & \textbf{6.701} & \textbf{2.385} & \textbf{92.1} & \textbf{0.023} & \textbf{6.723} & \textbf{2.382} & \textbf{91.4} & \underline{0.030} & \textbf{8.473} & 3.094 & \underline{84.7}\\
\hline
\end{tabular}
\label{tab:nsr_rgbdd}
\end{table*}

\begin{table*}[t]
\centering
\caption{The quantitative results comparison of the noisy depth super-resolution test on Middlebury 2014. The best are in \textbf{bold} and the second best are \underline{underlined} ones.}
\begin{tabular}{l@{ }c@{ }|c@{ }c@{ }c@{ }c@{ }|c@{ }c@{ }c@{ }c@{ }|c@{ }c@{ }c@{ }c}
\hline
\multirow{2}{*}{Method} & \multirow{2}{*}{reference} & \multicolumn{4}{c|}{4$\times$} & \multicolumn{4}{c|}{8$\times$} & \multicolumn{4}{c}{16$\times$}\\
 & & AbsRel$\downarrow$ & RMSE$\downarrow$ & iRMSE$\downarrow$ & $\delta_{1.05}\uparrow$ & AbsRel$\downarrow$ & RMSE$\downarrow$ & iRMSE$\downarrow$ & $\delta_{1.05}\uparrow$ & AbsRel$\downarrow$ & RMSE$\downarrow$ & iRMSE$\downarrow$ & $\delta_{1.05}\uparrow$ \\
\hline
DKN\cite{DKN2021} & IJCV2021 & 0.165 & 14.501 & 17.562 & 39.5 & 0.164 & 14.590 & 16.245 & 44.5 & 0.156 & 13.389 & 8.017 & 43.8\\
DCT\cite{DCTNet} & CVPR2022 & 0.177 & 13.647 & 5.703 & 15.6 & 0.181 & 13.640 & 6.252 & 18.8 & 0.188 & 13.578 & 7.780 & 20.9\\
DADA\cite{dada} & CVPR2023 & 0.155 & 14.015 & 11.747 & 43.7 & 0.160 & 14.358 & 15.370 & 44.1 & 0.180 & 15.034 & 20.565 & 38.8\\
C2F\cite{wang2023crf} & TIP2023 & 0.109 & \underline{7.765} & 6.260 & 49.0 & \underline{0.112} & \textbf{7.710} & 6.475 & 49.5 & \underline{0.122} & \textbf{8.169} & 6.098 & 45.2\\
G2\cite{wang2024g2} & PAMI2024 & \underline{0.100} & 13.050 & \underline{3.046} & \textbf{65.9} & 0.116 & 13.254 & {3.369} & \textbf{57.8} & 0.157 & 14.219 & \underline{5.164} & \underline{45.3}\\
SGN\cite{wang2024sgnet} & AAAI2024 & 0.141 & 13.396 & 5.041 & 44.7 & 0.145 & 13.445 & 5.013 & 44.1 & 0.156 & 13.261 & 5.615 & 40.5\\
Ours & - & \textbf{0.071} & \textbf{7.135} & \textbf{2.305} & \underline{55.9} & \textbf{0.075} & \underline{7.714} & \textbf{2.310} & \underline{57.4} & \textbf{0.092} & \underline{8.615} & \textbf{2.339} & \textbf{49.8}\\
\hline
\end{tabular}
\label{tab:nsr_mdb2014}
\end{table*}

\subsection{Ablation studies}
\label{ablation}
In this section, we conduct ablation studies to evaluate the effectiveness of each component in the proposed method. Specifically, we verify the effectiveness of the raw depth generation pipeline and the structure uncertainty module in recovering heavily distorted depth maps. Additionally, we verify the effectiveness of the feature alignment module across different major network backbones, the parameter update strategy, and related hyper-parameter settings in recovering raw depth captured by ToF sensors.

{\bf{Raw depth generation pipeline.}} The raw depth generation pipeline consists of two kinds of simulations designed to address various real-world raw depth conditions. One simulation focuses on traditional noisy low-resolution simulation (denoted as "NSR"), while the other aims the random structure misalignment (denoted as "RSM") to enrich the diversity of input raw depth maps. To comprehensively evaluate the pipeline, we establish four distinct conditions: (1) applying neither simulation, (2) applying the NSR simulation alone, (3) applying the RSM simulation alone, and (4) applying both simulations at the same time. The results in Table \ref{tab:raw_gen} demonstrate the effectiveness of each simulation, with the combination of both simulations achieving the best accuracy.



\begin{table}[!ht]      
\centering
\caption{Ablation study of raw depth generation pipeline.}
\label{tab:raw_gen}
\scalebox{1.00}{
\begin{tabular}{lc|cccc}
\hline            
NSR & RSM & AbsRel$\downarrow$ & RMSE$\downarrow$ & iRMSE$\downarrow$ & $\delta_{1.05}\uparrow$ \\ 
\hline  
 & & 0.108 & 13.096 & 3.654 & 57.7\\  
\checkmark & & 0.062 & 9.859 & 2.768 & 81.4\\  
 & \checkmark & 0.057 & 8.784 & 2.689 & 79.9\\  
\checkmark & \checkmark & \textbf{0.042} & \textbf{6.841} & \textbf{1.853} & \textbf{81.5}\\ 
\hline
\end{tabular}}
\end{table}

{\bf{The structure uncertainty module.}} To better adapt to various real-world raw depth maps, we design a structure uncertainty model to identify the challenging misaligned structures. In this module, we find that leveraging a depth foundation model (denoted as "DFM") \cite{midas, dav2} significantly enhances the measurement of raw structure uncertainty. We evaluate four configurations to demonstrate the effectiveness of the structure uncertainty module: (1) removing the structure uncertainty module, (2) adopting a simple structure uncertainty module without DFM, (3) using the structure uncertainty module with different DFMs (i.e., MiDaS \cite{midas} or DAV2 \cite{dav2}). Notably, DAV2, which is trained on larger mixed RGB-D datasets with more advanced techniques, outperforms MiDaS. The results in Table \ref{tab:uncertainty} demonstrate the effectiveness of the structure uncertainty module, with more advanced DFMs further improving the accuracy of structure uncertainty measurement.

Furthermore, the effectiveness of the structure uncertainty module can be extended to other recovery models, such as the comprehensive depth recovery model G2 \cite{wang2024g2}. By masking the estimated erroneous regions (setting them as 0) in raw depth maps, recovery methods achieve improved performance as they avoid preserving incorrect structures. Quantitative results demonstrating this improvement are presented in Table \ref{tab:uncertainty_app}. Without the structural uncertainty module, both G2 and our method exhibit limited capability in structure recovery. However, with the help of the proposed structure uncertainty module, the performance of both methods improves significantly, with our method achieving the best results.

\begin{table}[!ht]      
\centering
\caption{Ablation study of the structure uncertainty module.}
\label{tab:uncertainty}
\scalebox{1.00}{
\begin{tabular}{cc|cccc}
\hline            
Unc. & DFM & AbsRel$\downarrow$ & RMSE$\downarrow$ & iRMSE$\downarrow$ & $\delta_{1.05}\uparrow$ \\ 
\hline  
 & & 0.084 & 11.445 & 2.422 & 66.3\\ 
\checkmark & & 0.087 & 9.886 & 3.356 & 57.3\\  
\checkmark & MiDaS\cite{midas} & 0.068 & 9.431 & 2.103 & 77.4\\
\checkmark & DAV2\cite{dav2} & \textbf{0.042} & \textbf{6.841} & \textbf{1.853} & \textbf{81.5}\\ 
\hline
\end{tabular}}
\end{table}

\begin{table}[!ht]      
\centering
\caption{The effectiveness of structure uncertainty module in different models.}
\label{tab:uncertainty_app}
\scalebox{1.00}{
\begin{tabular}{cc|cccc}
\hline            
Method & Unc. & AbsRel$\downarrow$ & RMSE$\downarrow$ & iRMSE$\downarrow$ & $\delta_{1.05}\uparrow$ \\ 
\hline  
G2 \cite{wang2024g2} & & 0.085 & 13.153 & 2.602 & 73.5\\
G2 \cite{wang2024g2} & \checkmark & 0.062 & 9.378 & 1.988 & 73.7\\
Ours & & 0.084 & 11.445 & 2.422 & 66.3\\ 
Ours & \checkmark & \textbf{0.042} & \textbf{6.841} & \textbf{1.853} & \textbf{81.5}\\ 
\hline
\end{tabular}}
\end{table}

{\bf{The robust feature alignment module.}} We aim to design a robust module that effectively aligns the precise structure of RGB images while fitting real-world GT depth maps. The module is compatible with major network backbones, including U-Net \cite{wang2024g2}, Vision Transformer \cite{dav2}, and ConvNeXt \cite{woo2023convnext}, which is illustrated in Table \ref{tab:pap}. The results demonstrate that our feature alignment module (denoted as "FAM") enhances depth map recovery across different major backbones, with the ConvNeXt proved to be the optimal choice for the proposed method.

\begin{table}[!ht]      
\centering
\caption{Ablation study of the robust feature alignment module (FAM).}
\label{tab:pap}
\scalebox{1.00}{
\begin{tabular}{lc|cccc}
\hline            
architecture & FAM & AbsRel$\downarrow$ & RMSE$\downarrow$ & iRMSE$\downarrow$ & $\delta_{1.05}\uparrow$ \\ 
\hline  
U-Net\cite{wang2024g2} & & 0.016 & 5.388 & 1.920 & 95.8\\  
U-Net\cite{wang2024g2} & \checkmark & 0.015 & 5.249 & 1.871 & \textbf{96.0}\\  
ViT\cite{dav2} & & 0.018 & 5.800 & 2.079 & 95.0\\
ViT\cite{dav2} & \checkmark & \textbf{0.013} & 5.392 & 1.895 & 95.9\\
ConvNeXt\cite{woo2023convnext} & & 0.016 & 5.373 & 1.995 & 95.1\\
ConvNeXt\cite{woo2023convnext} & \checkmark & \textbf{0.013} & \textbf{5.109} & \textbf{1.836} & \textbf{96.0}\\
\hline
\end{tabular}}
\end{table}

Furthermore, to prevent overfitting to inaccurate GT depth in real-world scenarios, we adopt stochastic depth \cite{woo2023convnext} as the parameter update strategy in the feature alignment module. Alternative strategies, such as dropout \cite{dropout} and DropBlock \cite{dropblock}, can also be employed. We evaluate these strategies separately in Table \ref{tab:overfit}, and the results demonstrate the effectiveness of each parameter update strategy, with stochastic depth achieving the highest accuracy.

\begin{table}[!ht]      
\centering
\caption{Ablation study of avoiding overfitting strategy.}
\label{tab:overfit}
\scalebox{1.00}{
\begin{tabular}{l|cccc}
\hline            
strategy & AbsRel$\downarrow$ & RMSE$\downarrow$ & iRMSE$\downarrow$ & $\delta_{1.05}\uparrow$ \\ 
\hline  
- & 0.017 & 5.522 & 2.047 & 94.8\\  
Dropout\cite{dropout} & 0.014 & 5.195 & 1.853 & 95.6\\  
DropBlock\cite{dropblock} & 0.014 & 5.167 & \textbf{1.832} & 95.7\\
Stochastic depth\cite{woo2023convnext} & \textbf{0.013} & \textbf{5.109} & 1.836 & \textbf{96.0}\\
\hline
\end{tabular}}
\end{table}

We also evaluate the settings of hyper-parameters in the ablation study. First, we assess the impact of the distance $f$ between the query pixel and the target pixel in Table \ref{tab:para_f}, as described in Section \ref{sec:GDF}. Results in Table \ref{tab:para_f} indicate $f=5$ is the optimal choice in the proposed method. Besides, the channel dimension $C$ (the number of hidden layers) also affects the performance of the FAM in the proposed method. Results in Table \ref{tab:para_c} demonstrate $C=64$ yields the best performance in the proposed method. 

\begin{table}[!ht]      
\centering
\caption{Ablation study of hyper-parameter $f$.}
\label{tab:para_f}
\scalebox{1.00}{
\begin{tabular}{c|cccc}
\hline            
model & AbsRel$\downarrow$ & RMSE$\downarrow$ & iRMSE$\downarrow$ & $\delta_{1.05}\uparrow$ \\ 
\hline  
$f$=1 & 0.015 & 5.387 & 1.920 & 95.4\\  
$f$=3 & 0.015 & 5.374 & 1.929 & 95.3\\
$f$=5 & \textbf{0.013} & \textbf{5.109} & \textbf{1.836} & \textbf{96.0}\\
$f$=7 & 0.016 & 5.360 & 1.986 & 95.1\\  
\hline
\end{tabular}}
\end{table}

\begin{table}[!ht]      
\centering
\caption{Ablation study of hyper-parameter $C$.}
\label{tab:para_c}
\scalebox{1.00}{
\begin{tabular}{c|cccc}
\hline            
model & AbsRel$\downarrow$ & RMSE$\downarrow$ & iRMSE$\downarrow$ & $\delta_{1.05}\uparrow$ \\ 
\hline  
$C$=32 & 0.014 & 5.231 & 1.874 & 95.6\\  
$C$=64 & \textbf{0.013} & \textbf{5.109} & \textbf{1.836} & \textbf{96.0}\\
$C$=96 & 0.017 & 5.521 & 2.047 & 94.9\\
$C$=128 & 0.018 & 5.535 & 2.080 & 94.7\\
\hline
\end{tabular}}
\end{table}

\section{Conclusion}
\label{sec5}

In this paper, we propose a robust RGB-guided depth recovery framework to recover various raw depth maps in the real world. The ability of structure correction in the proposed method mainly comes from the considering of random structure misalignments for both input raw depth maps and output depth maps. For input, a raw depth generation pipeline is designed to address diverse structure misalignments rather than handling raw depth in a specific condition. Additionally, a structure uncertainty module is proposed to identify the challenging misaligned structure, helping the recovery model better generalize in unseen scenarios. For output, we introduce a robust feature alignment module, which aligns with the precise structure of RGB images while mitigating overfitting to inaccurate GT depth maps. This module demonstrates strong compatibility with major network backbones, particularly ConvNeXt. Extensive experiments on multiple datasets demonstrate that the proposed method achieves competitive accuracy and generalization capabilities across various challenging raw depth maps, outperforming recent RGB-guided depth recovery methods in both qualitative and quantitative evaluations. 

\bibliographystyle{IEEEtran}

\begin{thebibliography}{10}
\providecommand{\url}[1]{#1}
\csname url@samestyle\endcsname
\providecommand{\newblock}{\relax}
\providecommand{\bibinfo}[2]{#2}
\providecommand{\BIBentrySTDinterwordspacing}{\spaceskip=0pt\relax}
\providecommand{\BIBentryALTinterwordstretchfactor}{4}
\providecommand{\BIBentryALTinterwordspacing}{\spaceskip=\fontdimen2\font plus
\BIBentryALTinterwordstretchfactor\fontdimen3\font minus \fontdimen4\font\relax}
\providecommand{\BIBforeignlanguage}[2]{{%
\expandafter\ifx\csname l@#1\endcsname\relax
\typeout{** WARNING: IEEEtran.bst: No hyphenation pattern has been}%
\typeout{** loaded for the language `#1'. Using the pattern for}%
\typeout{** the default language instead.}%
\else
\language=\csname l@#1\endcsname
\fi
#2}}
\providecommand{\BIBdecl}{\relax}
\BIBdecl

\bibitem{cong2024gradient}
W.~Cong, Y.~Cong, J.~Dong, G.~Sun, and H.~Ding, ``Gradient-semantic compensation for incremental semantic segmentation,'' \emph{IEEE Transactions on Multimedia}, vol.~26, pp. 5561 -- 5574, 2024.

\bibitem{gao2023dasi}
J.~Gao, D.~Kong, S.~Wang, J.~Li, and B.~Yin, ``Dasi: Learning domain adaptive shape impression for 3d object reconstruction,'' \emph{IEEE Transactions on Multimedia}, vol.~25, pp. 5248 -- 5262, 2022.

\bibitem{RenDongRan}
D.~Ren, M.~Yang, J.~Wu, and N.~Zheng, ``Surface normal and gaussian weight constraints for indoor depth structure completion,'' \emph{Pattern Recognition}, vol. 138, p. 109362, 2023.

\bibitem{PRFANG2023109139}
X.~Fang, M.~Jiang, J.~Zhu, X.~Shao, and H.~Wang, ``M2rnet: Multi-modal and multi-scale refined network for rgb-d salient object detection,'' \emph{Pattern Recognition}, vol. 135, p. 109139, 2023.

\bibitem{chun2024usd}
D.~Chun, S.~Lee, and H.~Kim, ``Usd: Uncertainty-based one-phase learning to enhance pseudo-label reliability for semi-supervised object detection,'' \emph{IEEE Transactions on Multimedia}, vol.~26, pp. 6336 -- 6347, 2024.

\bibitem{agrawal2022dc}
A.~Agrawal, S.~Hariharan, A.~S. Bedi, and D.~Manocha, ``Dc-mrta: Decentralized multi-robot task allocation and navigation in complex environments,'' in \emph{Proceeding of the IEEE/RSJ International Conference on Intelligent Robots and Systems (IROS)}.\hskip 1em plus 0.5em minus 0.4em\relax IEEE, 2022, pp. 11\,711--11\,718.

\bibitem{nyuv2}
N.~Silberman, D.~Hoiem, P.~Kohli, and R.~Fergus, ``Indoor segmentation and support inference from rgbd images,'' in \emph{Proceedings of the European Conference on Computer Vision (ECCV)}.\hskip 1em plus 0.5em minus 0.4em\relax Springer, 2012, pp. 746--760.

\bibitem{ibims}
T.~Koch, L.~Liebel, F.~Fraundorfer, and M.~Korner, ``Evaluation of cnn-based single-image depth estimation methods,'' in \emph{Proceedings of the European Conference on Computer Vision (ECCV) Workshops}, 2018, pp. 331--348.

\bibitem{li2018megadepth}
Z.~Li and N.~Snavely, ``Megadepth: Learning single-view depth prediction from internet photos,'' in \emph{Proceedings of the IEEE conference on computer vision and pattern recognition}, 2018, pp. 2041--2050.

\bibitem{midas}
R.~Ranftl, K.~Lasinger, D.~Hafner, K.~Schindler, and V.~Koltun, ``Towards robust monocular depth estimation: Mixing datasets for zero-shot cross-dataset transfer,'' \emph{IEEE transactions on pattern analysis and machine intelligence}, vol.~44, no.~3, pp. 1623--1637, 2022.

\bibitem{GuidedFilter}
K.~He, J.~Sun, and X.~Tang, ``Guided image filtering,'' \emph{IEEE Transactions on Pattern Analysis and Machine Intelligence}, vol.~35, no.~6, pp. 1397--1409, 2012.

\bibitem{zuo2016mrf}
Y.~Zuo, Q.~Wu, J.~Zhang, and P.~An, ``Explicit edge inconsistency evaluation model for color-guided depth map enhancement,'' \emph{IEEE Transactions on Circuits and Systems for Video Technology}, vol.~28, no.~2, pp. 439--453, 2016.

\bibitem{wang2023crf}
H.~Wang, M.~Yang, C.~Zhu, and N.~Zheng, ``Rgb-guided depth map recovery by two-stage coarse-to-fine dense crf models,'' \emph{IEEE Transactions on Image Processing}, vol.~32, pp. 1315--1328, 2023.

\bibitem{DKN2021}
B.~Kim, J.~Ponce, and B.~Ham, ``Deformable kernel networks for joint image filtering,'' \emph{International Journal of Computer Vision}, vol. 129, no.~2, pp. 579--600, 2021.

\bibitem{DAGF}
Z.~Zhong, X.~Liu, J.~Jiang, D.~Zhao, and X.~Ji, ``Deep attentional guided image filtering,'' \emph{IEEE Transactions on Neural Networks and Learning Systems}, 2023.

\bibitem{wang2024sgnet}
Z.~Wang, Z.~Yan, and J.~Yang, ``Sgnet: Structure guided network via gradient-frequency awareness for depth map super-resolution,'' in \emph{Proceedings of the AAAI Conference on Artificial Intelligence}, vol.~38, no.~6, 2024, pp. 5823--5831.

\bibitem{he2021towards}
L.~He, H.~Zhu, F.~Li, H.~Bai, R.~Cong, C.~Zhang, C.~Lin, M.~Liu, and Y.~Zhao, ``Towards fast and accurate real-world depth super-resolution: Benchmark dataset and baseline,'' in \emph{Proceedings of the ieee/cvf conference on computer vision and pattern recognition}, 2021, pp. 9229--9238.

\bibitem{DCTNet}
Z.~Zhao, J.~Zhang, S.~Xu, Z.~Lin, and H.~Pfister, ``Discrete cosine transform network for guided depth map super-resolution,'' in \emph{Proceedings of the IEEE/CVF Conference on Computer Vision and Pattern Recognition}, 2022, pp. 5697--5707.

\bibitem{yan2018ddrnet}
S.~Yan, C.~Wu, L.~Wang, F.~Xu, L.~An, K.~Guo, and Y.~Liu, ``Ddrnet: Depth map denoising and refinement for consumer depth cameras using cascaded cnns,'' in \emph{Proceedings of the European conference on computer vision (ECCV)}, 2018, pp. 151--167.

\bibitem{wang2023self}
J.~Wang, P.~Liu, and F.~Wen, ``Self-supervised learning for rgb-guided depth enhancement by exploiting the dependency between rgb and depth,'' \emph{IEEE Transactions on Image Processing}, vol.~32, pp. 159--174, 2023.

\bibitem{dada}
N.~Metzger, R.~C. Daudt, and K.~Schindler, ``Guided depth super-resolution by deep anisotropic diffusion,'' in \emph{Proceedings of the IEEE/CVF Conference on Computer Vision and Pattern Recognition (CVPR)}, 2023, pp. 18\,237--18\,246.

\bibitem{qiao2023self}
X.~Qiao, C.~Ge, C.~Zhao, F.~Tosi, M.~Poggi, and S.~Mattoccia, ``Self-supervised depth super-resolution with contrastive multiview pre-training,'' \emph{Neural Networks}, vol. 168, pp. 223--236, 2023.

\bibitem{simard2003best}
P.~Y. Simard, D.~Steinkraus, J.~C. Platt \emph{et~al.}, ``Best practices for convolutional neural networks applied to visual document analysis.'' in \emph{Icdar}, vol.~3, no. 2003.\hskip 1em plus 0.5em minus 0.4em\relax Edinburgh, 2003.

\bibitem{touvron2022resmlp}
H.~Touvron, P.~Bojanowski, M.~Caron, M.~Cord, A.~El-Nouby, E.~Grave, G.~Izacard, A.~Joulin, G.~Synnaeve, J.~Verbeek \emph{et~al.}, ``Resmlp: Feedforward networks for image classification with data-efficient training,'' \emph{IEEE transactions on pattern analysis and machine intelligence}, vol.~45, no.~4, pp. 5314--5321, 2022.

\bibitem{dav2}
L.~Yang, B.~Kang, Z.~Huang, Z.~Zhao, X.~Xu, J.~Feng, and H.~Zhao, ``Depth anything {V2},'' in \emph{Advances in Neural Information Processing Systems 38: Annual Conference on Neural Information Processing Systems 2024, NeurIPS 2024, Vancouver, BC, Canada, December 10 - 15, 2024}, 2024.

\bibitem{liif}
Y.~Chen, S.~Liu, and X.~Wang, ``Learning continuous image representation with local implicit image function,'' in \emph{Proceedings of the IEEE/CVF conference on computer vision and pattern recognition}, 2021, pp. 8628--8638.

\bibitem{wang2024g2}
H.~Wang, M.~Yang, and N.~Zheng, ``G2-monodepth: A general framework of generalized depth inference from monocular rgb+x data,'' \emph{IEEE Transactions on Pattern Analysis and Machine Intelligence}, vol.~46, no.~5, pp. 3753--3771, 2024.

\bibitem{woo2023convnext}
S.~Woo, S.~Debnath, R.~Hu, X.~Chen, Z.~Liu, I.~S. Kweon, and S.~Xie, ``Convnext v2: Co-designing and scaling convnets with masked autoencoders,'' in \emph{Proceedings of the IEEE/CVF Conference on Computer Vision and Pattern Recognition}, 2023, pp. 16\,133--16\,142.

\bibitem{rgbdd}
L.~He, H.~Zhu, F.~Li, H.~Bai, R.~Cong, C.~Zhang, C.~Lin, M.~Liu, and Y.~Zhao, ``Towards fast and accurate real-world depth super-resolution: Benchmark dataset and baseline,'' in \emph{Proceedings of the IEEE/CVF Conference on Computer Vision and Pattern Recognition}, 2021, pp. 9229--9238.

\bibitem{middlebury2014}
D.~Scharstein, H.~Hirschm{\"u}ller, Y.~Kitajima, G.~Krathwohl, N.~Ne{\v{s}}i{\'c}, X.~Wang, and P.~Westling, ``High-resolution stereo datasets with subpixel-accurate ground truth,'' in \emph{Proceedings of the 36th German Conference on Pattern Recognition, M{\"u}nster, Germany, September 2-5, 2014}.\hskip 1em plus 0.5em minus 0.4em\relax Springer, 2014, pp. 31--42.

\bibitem{sgbm}
H.~Hirschmuller, ``Stereo processing by semiglobal matching and mutual information,'' \emph{IEEE Transactions on pattern analysis and machine intelligence}, vol.~30, no.~2, pp. 328--341, 2007.

\bibitem{bf}
Y.~Wang, A.~Ortega, D.~Tian, and A.~Vetro, ``A graph-based joint bilateral approach for depth enhancement,'' in \emph{2014 IEEE International Conference on Acoustics, Speech and Signal Processing (ICASSP)}.\hskip 1em plus 0.5em minus 0.4em\relax IEEE, 2014, pp. 885--889.

\bibitem{zhang2014new}
P.~Zhang and F.~Li, ``A new adaptive weighted mean filter for removing salt-and-pepper noise,'' \emph{IEEE signal processing letters}, vol.~21, no.~10, pp. 1280--1283, 2014.

\bibitem{Ham2017RobustGIF}
B.~Ham, M.~Cho, and J.~Ponce, ``Robust guided image filtering using nonconvex potentials,'' \emph{IEEE Transactions on Pattern Analysis and Machine Intelligence}, vol.~40, no.~1, pp. 192--207, 2017.

\bibitem{liu2021generalized}
W.~Liu, P.~Zhang, Y.~Lei, X.~Huang, J.~Yang, and M.~Ng, ``A generalized framework for edge-preserving and structure-preserving image smoothing,'' \emph{IEEE Transactions on Pattern Analysis and Machine Intelligence}, vol.~44, no.~10, pp. 6631--6648, 2021.

\bibitem{zuo2018minimum}
Y.~Zuo, Q.~Wu, J.~Zhang, and P.~An, ``Minimum spanning forest with embedded edge inconsistency measurement model for guided depth map enhancement,'' \emph{IEEE Transactions on Image Processing}, vol.~27, no.~8, pp. 4145--4159, 2018.

\bibitem{liu2016robust}
W.~Liu, X.~Chen, J.~Yang, and Q.~Wu, ``Robust color guided depth map restoration,'' \emph{IEEE Transactions on Image Processing}, vol.~26, no.~1, pp. 315--327, 2016.

\bibitem{wang2022tip}
H.~Wang, M.~Yang, X.~Lan, C.~Zhu, and N.~Zheng, ``Depth map recovery based on a unified depth boundary distortion model,'' \emph{IEEE Transactions on Image Processing}, vol.~31, pp. 7020--7035, 2022.

\bibitem{zuo2019depth}
Y.~Zuo, Y.~Fang, Y.~Yang, X.~Shang, and Q.~Wu, ``Depth map enhancement by revisiting multi-scale intensity guidance within coarse-to-fine stages,'' \emph{IEEE Transactions on Circuits and Systems for Video Technology}, vol.~30, no.~12, pp. 4676--4687, 2019.

\bibitem{yan2024learning}
Z.~Yan, K.~Wang, X.~Li, Z.~Zhang, G.~Li, J.~Li, and J.~Yang, ``Learning complementary correlations for depth super-resolution with incomplete data in real world,'' \emph{IEEE transactions on neural networks and learning systems}, vol.~35, no.~4, pp. 5616--5626, 2024.

\bibitem{kitti}
A.~Geiger, P.~Lenz, and R.~Urtasun, ``Are we ready for autonomous driving? the kitti vision benchmark suite,'' in \emph{Proceedings of the IEEE conference on Computer Vision and Pattern Recognition (CVPR)}.\hskip 1em plus 0.5em minus 0.4em\relax IEEE, 2012, pp. 3354--3361.

\bibitem{matterport3d}
A.~Chang, A.~Dai, T.~Funkhouser, M.~Halber, M.~Niessner, M.~Savva, S.~Song, A.~Zeng, and Y.~Zhang, ``Matterport3d: Learning from rgb-d data in indoor environments,'' \emph{arXiv preprint arXiv:1709.06158}, 2017.

\bibitem{vkitti}
A.~Gaidon, Q.~Wang, Y.~Cabon, and E.~Vig, ``Virtual worlds as proxy for multi-object tracking analysis,'' in \emph{Proceedings of the IEEE conference on Computer Vision and Pattern Recognition (CVPR)}, 2016, pp. 4340--4349.

\bibitem{kirillov2023segment}
A.~Kirillov, E.~Mintun, N.~Ravi, H.~Mao, C.~Rolland, L.~Gustafson, T.~Xiao, S.~Whitehead, A.~C. Berg, W.-Y. Lo \emph{et~al.}, ``Segment anything,'' in \emph{Proceedings of the IEEE/CVF international conference on computer vision}, 2023, pp. 4015--4026.

\bibitem{unet}
O.~Ronneberger, P.~Fischer, and T.~Brox, ``U-net: Convolutional networks for biomedical image segmentation,'' in \emph{Medical image computing and computer-assisted intervention--MICCAI 2015: 18th international conference, Munich, Germany, October 5-9, 2015, proceedings, part III 18}.\hskip 1em plus 0.5em minus 0.4em\relax Springer, 2015, pp. 234--241.

\bibitem{dropout}
K.~KC, R.~Li, and M.~Gilany, ``Joint inference for neural network depth and dropout regularization,'' \emph{Advances in neural information processing systems}, vol.~34, pp. 26\,622--26\,634, 2021.

\bibitem{dropblock}
J.~Yao, W.~Xing, D.~Wang, J.~Xing, and L.~Wang, ``Active dropblock: Method to enhance deep model accuracy and robustness,'' \emph{Neurocomputing}, vol. 454, pp. 189--200, 2021.

\bibitem{xian2020structure}
K.~Xian, J.~Zhang, O.~Wang, L.~Mai, Z.~Lin, and Z.~Cao, ``Structure-guided ranking loss for single image depth prediction,'' in \emph{Proceedings of the IEEE/CVF conference on Computer Vision and Pattern Recognition (CVPR)}, 2020, pp. 611--620.

\bibitem{hu2022deep}
J.~Hu, C.~Bao, M.~Ozay, C.~Fan, Q.~Gao, H.~Liu, and T.~L. Lam, ``Deep depth completion from extremely sparse data: A survey,'' \emph{IEEE Transactions on Pattern Analysis and Machine Intelligence}, vol.~45, no.~7, pp. 8244--8264, 2022.

\end{thebibliography}

\end{document}